# Digital Elevation Model enhancement using Deep Learning

Casey Handmer, NASA Jet Propulsion Laboratory, California Institute of Technology, casey.j.handmer@jpl.nasa.gov
MC 138-204, 4800 Oak Grove Dr, Pasadena, CA 91109

## Abstract

We demonstrate high fidelity enhancement of planetary digital elevation models (DEMs) using optical images and deep learning with convolutional neural networks. Enhancement can be applied recursively to the limit of available optical data, representing a 90x resolution improvement in global Mars DEMs. Deep learning-based photoclinometry robustly recovers features obscured by non-ideal lighting conditions. Method can be automated at global scale. Analysis shows enhanced DEM slope errors are comparable with high resolution maps using conventional, labor intensive methods.

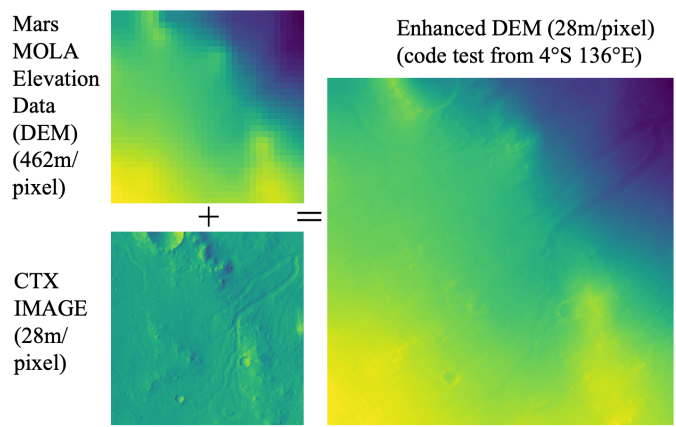

Figure 1. Conceptual summary of CTX-guided MOLA enhancement by neural network.

## Introduction

The Mars Global Surveyor (MGS) instrument Mars Orbital Laser Altimeter (MOLA) produced a global 512 ppd DEM, with a resolution of 462 m/pixel[1]. Mars Reconnaissance Orbiter (MRO) instrument Context Imager (CTX) has imaged >99.9% of the planet at 6 m resolution[2]. These images have been collated into an incredibly useful global dataset by the Murray Lab[3].

Historically, optical images of Mars have been converted into DEMs using stereophotogrammetry[4] and photoclinometry (shape from shading)[5]. Stereophotogrammetry can suffer from noise due to low visual contrast surfaces, while conventional photoclinometry is non-unique, non-absolute, and tends to lack detail on slopes parallel to the illumination direction[6]. Both methods are highly develop. For example, photoclinometric work on Mars today involves sophisticated radiance modeling to incorporate the effects of multiple point light scattering and atmospheric distortion[7].

Deep learning is an emerging machine learning (ML) mechanism which has proven to be successful in problems ranging from big data prediction[8] to classification[9] to game play[10]. The classic demo "pix2pix" showcases the ability of convolutional neural networks (CNNs) to transform images to other images. Example applications include image enhancement[11], historical image colorization[12], and increasingly sophisticated deep fakes[13].

In particular, pix2pix-like systems are quite adept at learning how to invert lossy processes. In our case, producing a radiometrically realistic image from DEM, albedo, and illumination data is fairly straightforward. Graphics kernels such as Unity[14] or Unreal Engine[15] do this hundreds of times a second in computer games. The inverse problem is difficult in a classical sense, as rendering is lossy, noisy, and the inverse is not uniquely determined. Despite this, pix2pix internalizes complex priors, collapsing the potential inverse space enough to produce plausible solutions.

We demonstrate an implementation of pix2pix that uses high resolution optical imagery to robustly enhance a low resolution DEM on Mars.

## Methods

Our Optical DEM Enhancement convolutional neural Network (ODENet) consists of two separate convolutional networks, trained sequentially.

The first upsamples a 16x16 pixel MOLA DEM sample to 32x32, and is trained on MOLA-derived DEM data. This "InterpNet" tends not to add the high spatial frequency data lost in downsampling.

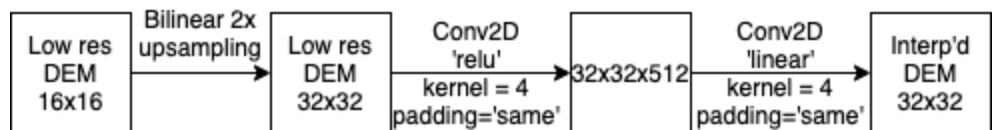

Figure 2. 2x upsampling performed by basic naive interpolator.

There is, therefore, a residual between the InterpNet result and the underlying ground truth. This residual, composed almost exclusively of high spatial frequency detail, needs to be recovered for useful DEM enhancement, especially when applying a 2x resolution enhancement recursively.

The second CNN ("EnhanceNet") uses a pix2pix-derived algorithm to convert a 32x32 pixel visual CTX image into the InterpNet residual. Since the raw CTX dataset has about 90 times higher resolution than MOLA, it has no shortage of high spatial frequency information to exploit.

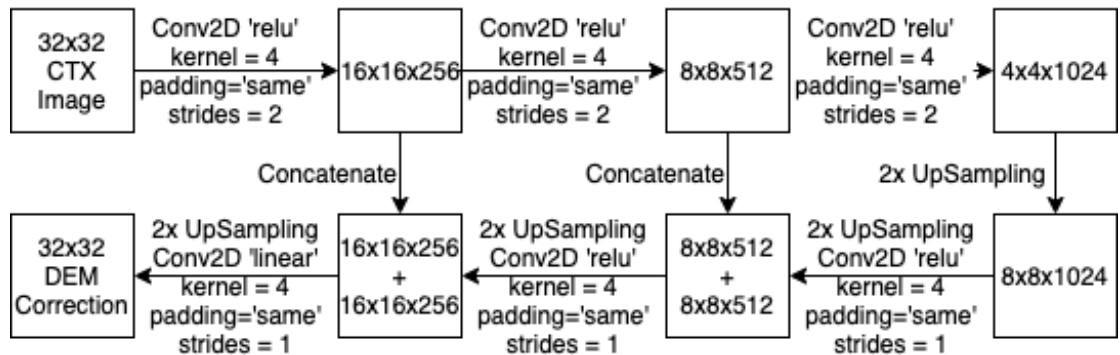

Figure 3. Upsampler error correction performed by CTX-derived high spatial frequency neural network.

Both networks can train in less than an hour on a standard MacBook Pro. Code and sample models are available on request.

Together these two networks form the Optical Digital elevation model Enhancement Network (ODENet), a function that takes a 16x16 DEM and 32x32 image of the same area and produces a 32x32 DEM, a resolution improvement of a factor of two.

ODENet is typically more accurate away from the edge of the 32x32 tile, so may be employed as a sliding window over larger datasets. It may be applied recursively to any desired level of resolution enhancement, limited only by optical image resolution.

## Validation

Deep learning algorithms such as pix2pix are recognized as being highly capable of producing plausible-looking data, in many cases good enough to fool even expert judges. That isn't necessarily good enough for the science. We validate the performance of ODENet in two ways: direct comparison with higher resolution data and directional power spectrum analysis.

### Comparison with HRSC dataset

We performed a 4x resolution enhancement on the Gale Crater quad (-8 - -4 N, 136 – 140 E) MOLA DEM, and compared it to HRSC-derived 100m resolution DEM data on an overlapping area[16]. We compared absolute elevations on points corresponding to the original MOLA dataset, the 2x, and the 4x enhancement.

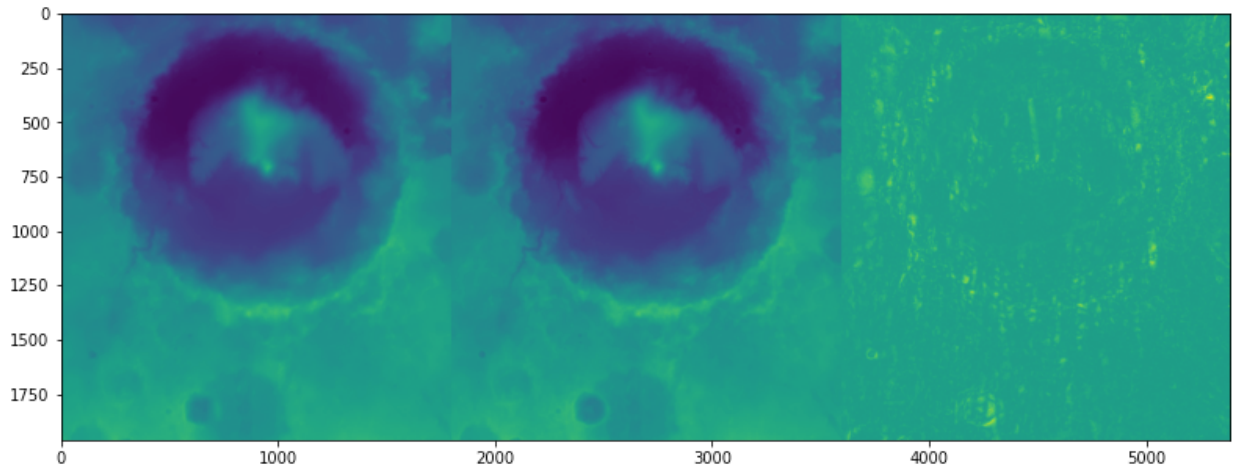

Figure 4. 4x enhanced MOLA DEM (left), HRSC (center), and difference (right) show that, once a lat/lon registration mismatch of (0.0139E, 0.00284N) is corrected, DEM differences are dominated by north-south oriented features that appear to be the ground tracks of various satellites. This implies that enhancement numerical errors are small compared with the intrinsic systematic errors typical of this sort of dataset.

As errors accumulate with recursive enhancement, slopes are a scale-independent way of assessing accuracy.

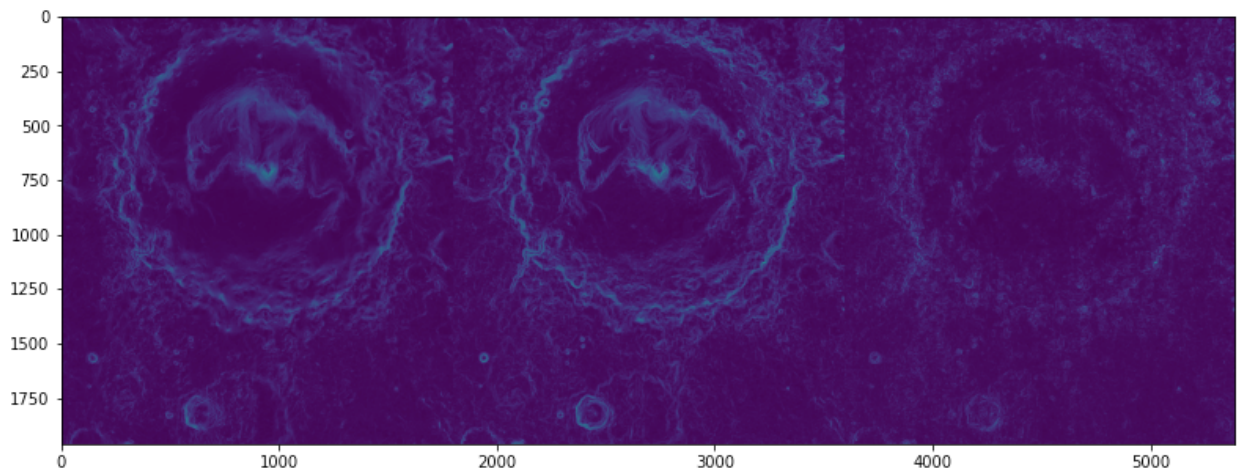

Figure 5. Slopes of 4x MOLA enhancement (left), HRSC dataset (center), and vector error (right) show qualitative agreement.

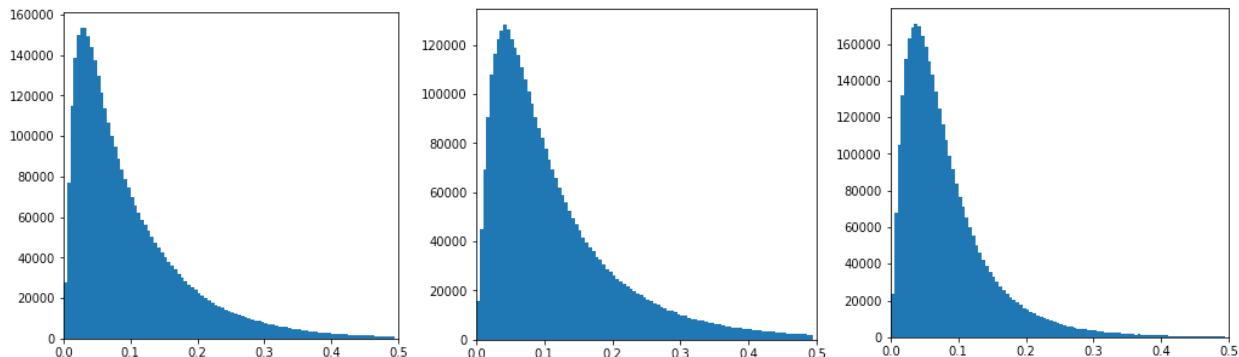

Figure 6. Slope gradient histograms of 4x enhancement (left), HRSC data (center), and vector slope error (right) show that most of the Gale quad is close to flat, and most of the slope errors are small variations on flatter terrain.

We found that the absolute error was within the expected range for systematic error in DEM data at this resolution, and that slope error was consistent with both absolute DEM systematic error and any cumulative error gained during recursive enhancement.

### Directional power spectrum analysis

A common weakness of photoclinometric DEMs is a lack of detail in features parallel to the illumination direction. This can be quantified by examining the 2D discrete Fourier transform of the DEM array, which can lack power in this direction. The 2D Fourier transform of the Gale quad is examined in detail below. ODENet learns to compensate for this low shading contrast on slopes facing away from the illumination direction.

A common issue with Fourier-based power analysis is low frequency "star burst" patterns due to non-periodic boundaries of data arrays. To avoid this, we blended the array edges with a Gaussian mask, as shown in Figure 7 and 8.

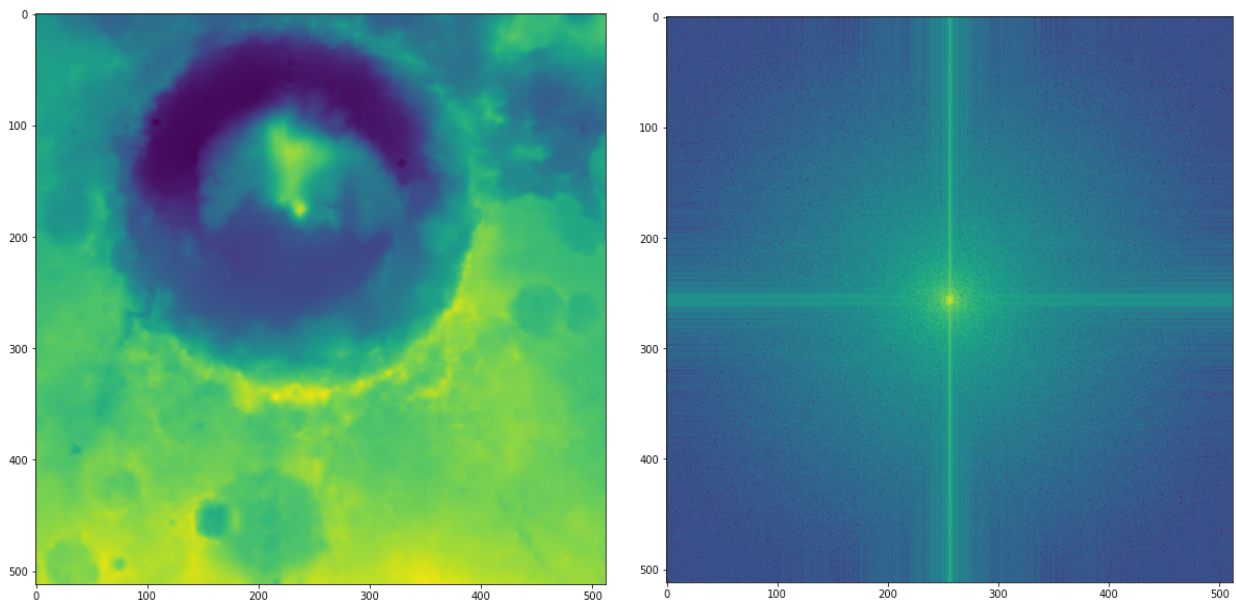

Figure 7. MOLA DEM and FFT showing low frequency starburst due to non-periodic data.

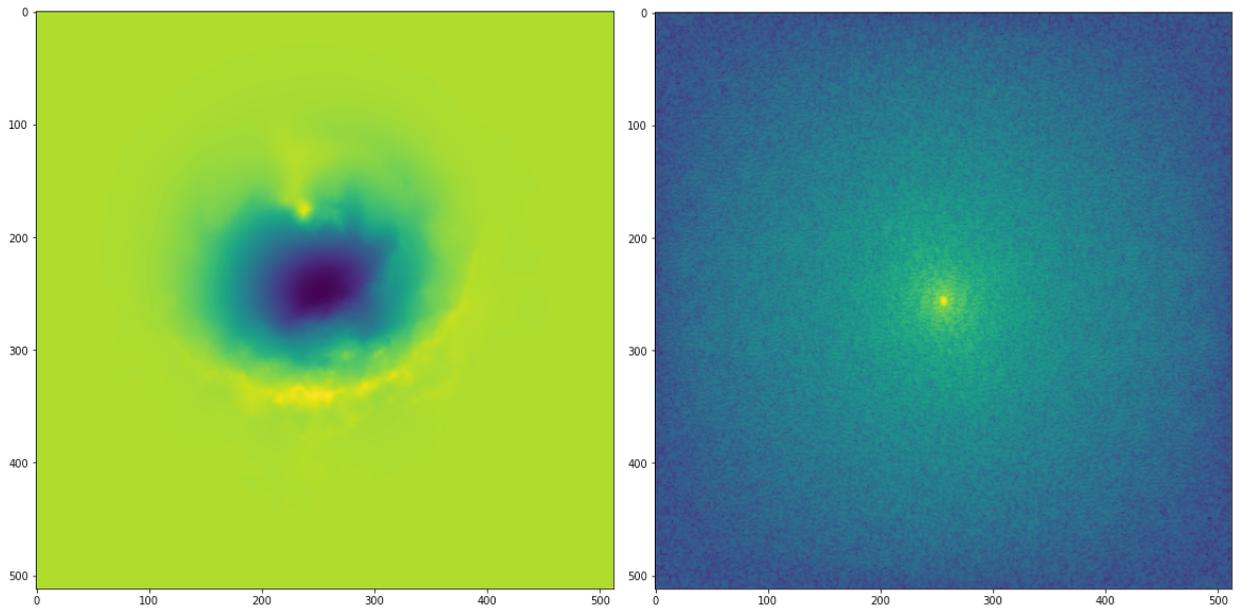

Figure 8. Gaussian-blended MOLA DEM and FFT showing reduced low frequency noise and unaffected high frequency power.

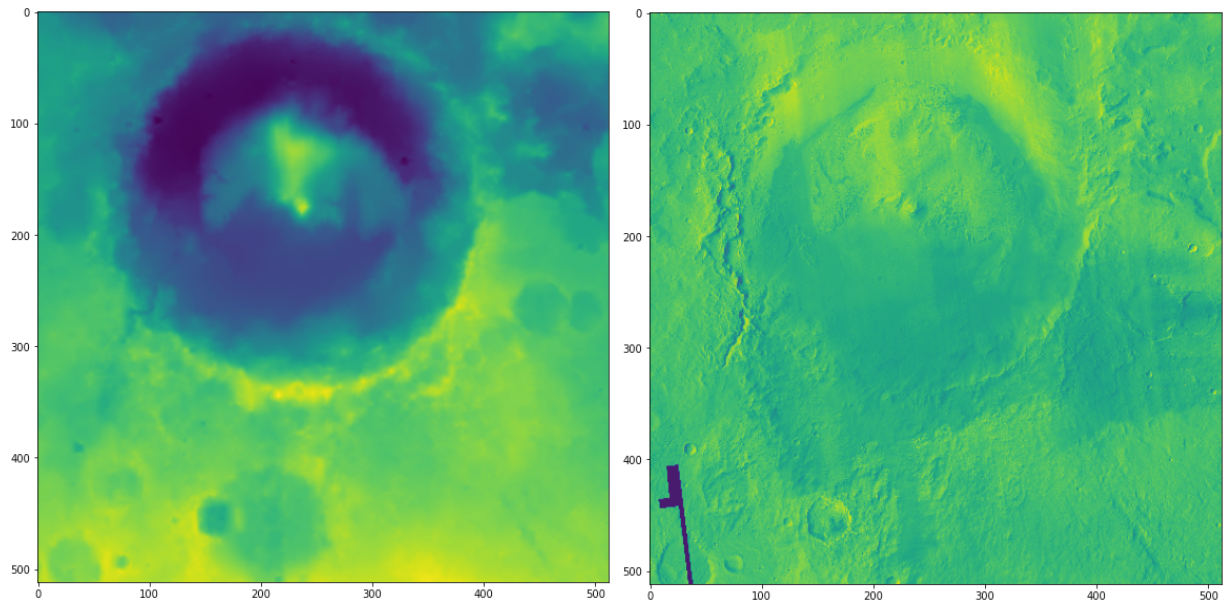

Figure 9. MOLA (left) and CTX-derived (right) images of the Gale crater quad at MOLA resolution.

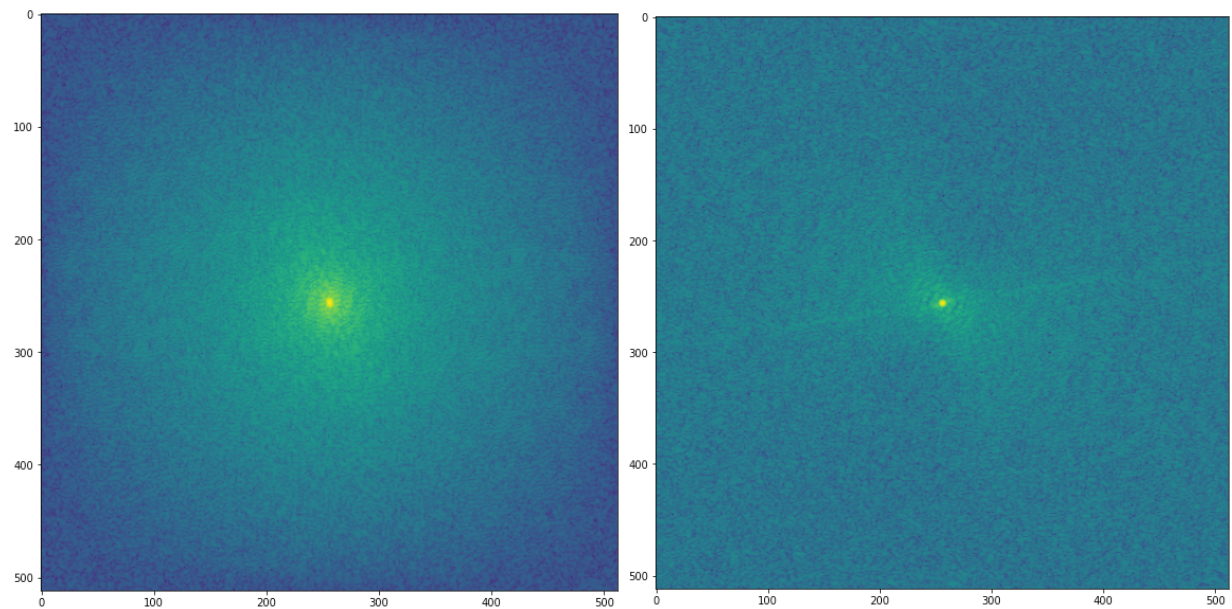

Figure 10. 2D FFTs (log scale of absolute value) of the above images show an equal power spectral density in the case of the MOLA DEM (left) and visually salient depletion of power on the 1 o'clock and 7 o'clock position in the CTX FFT (right). Artifacts in the right image are due to missing CTX data in the bottom left of the image.

The question is how to make this power depletion quantitative. We interpolated the FFTs to generate radial transects, enabling a numerical polar transform.

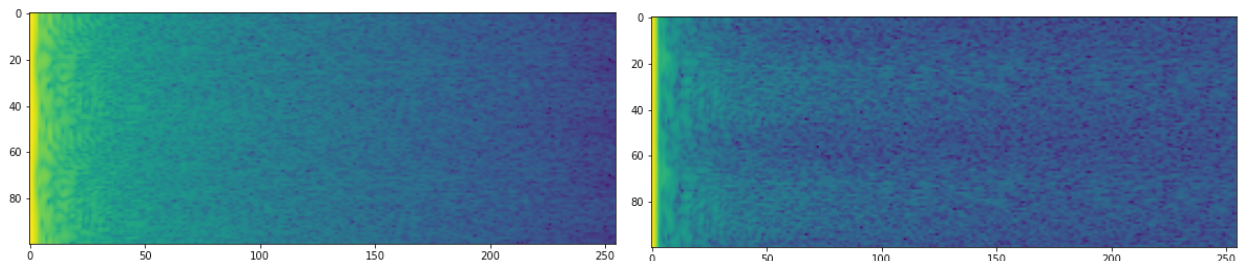

Figure 11. Polar transform of MOLA FFT (left) and CTX FFT (right), showing that the CTX image has noticeably more power, as well as angular depletion, at high frequencies.

The relatively steep decline in spectral power seen in the polar MOLA FFT compared to the CTX FFT in Figure 11 is a corollary of their differing fractal dimension. Not only does MOLA not resolve small features very well, their amplitude is roughly proportional to their size. In contrast, optical images have scale-invariant transitions between light and shadow, increasing the relative power at high frequencies. This is fortuitous as it renders the power deficit in the transverse solar direction more obvious.

Next, we marginalize the polar FFTs by summing over the radial coordinate.

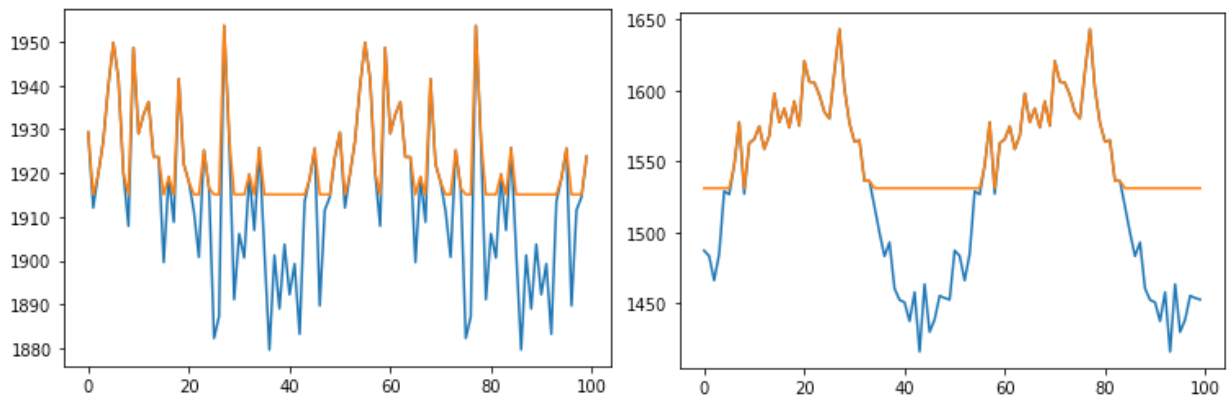

Figure 12. The marginalized MOLA polar FFT (left) shows no sinusoidal power variation, while the CTX case (right) does. The orange curve is pegged to the mean value, giving a measure of image spectral power deficit below that level.

To reduce this to a single number, we once again take a Fourier transform and compare the amplitude for the two-peaked periodic term to the constant (DC) term. For MOLA, this is 0.0037, while for CTX it is 0.027. These numbers can be thought of as a photoclinometry accuracy score, where a number closer to the MOLA baseline is more representative of a DEM that includes detail relatively obscured by a lack of shading in the illumination direction.

How does this look for the enhanced DEMs?

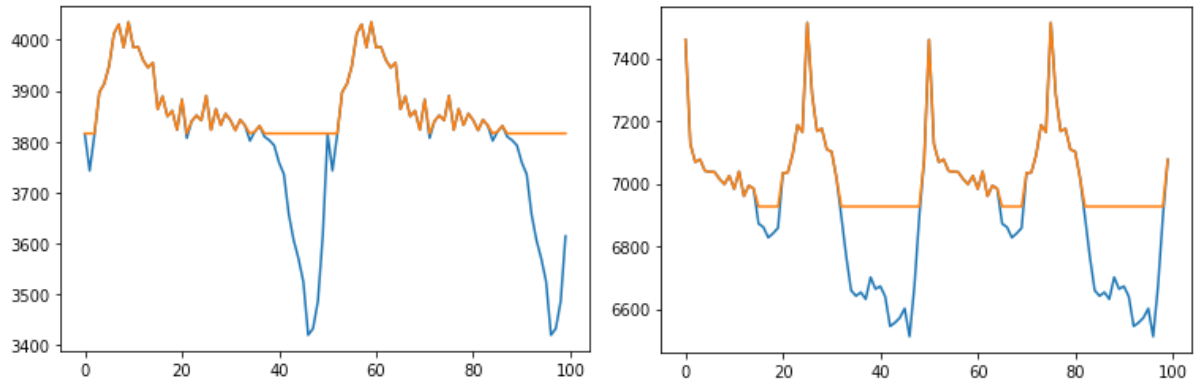

Figure 13. ODENet MOLA enhancements' marginalized polar FFT power at 2x enhancement (left) and 4x enhancement (right). This gives a measure of angular power depletion due to photoclinometric limitations.

Figure 13 shows a moderate spectral power deficit corresponding to the illumination direction. Quantitatively, they achieve a score of 0.018 and 0.013 respectively. Practically speaking, the ODENet corrector has recovered around 50% of the off-angle high spatial frequency DEM signal, which is otherwise obscured by unfavorable lighting.

These scores can be used to benchmark enhancement performance when comparing different algorithms, locations, and degree of enhancement.

ODENet has comparable accuracy to existing, labor-intensive high resolution DEM production methods, inviting its application to global datasets.

## Results

In addition to the 4x enhancement discussed in the validation section, the automated utility of ODENet enables the production and release of other data products potentially of interest to researchers.

A global 115 m resolution Mars DEM and select quads at 14 m resolution are now available by request.

## Discussion

ODENet-derived extraction of high quality, high resolution DEM data from existing lower resolution DEMs and high resolution optical data is a powerful technique. By open sourcing the underlying code, we intend to enable other researchers to refine and extend the method to LOLA/LROC, New Horizons, Europa Clipper, and MESSENGER data.

We also anticipate that the method may be applicable to optical pix2pix enhancement of other datasets, such as albedo and thermal properties.

Paper draft – Released for External Review Jan 12 2021 (URS 297718)

## Acknowledgements
Author would like to acknowledge Hiro Ono and Fred Calef for helpful discussions and assistance during formulation of quantitative performance evaluation. Also, Jay Dickson for helpful discussions and the vision behind the Murray Lab CTX dataset. Partly funded by MAARS (01STCR - R.18.239.014).

This research was carried out at the Jet Propulsion Laboratory, California Institute of Technology, under a contract with the National Aeronautics and Space Administration.

Reference herein to any specific commercial product, process, or service by trade name, trademark, manufacturer, or otherwise, does not constitute or imply its endorsement by the United States Government or the Jet Propulsion Laboratory, California Institute of Technology.

## References

[1] Smith, D.E., Zuber, M.T., Neumann, G.A., Guinness, E.A., Slavney, S., 2003. Mars Global Surveyor Laser Altimeter Mission Experiment Gridded Data Record, MGS-M-MOLA- 5-MEGDR-L3-V1.0, NASA Planetary Data System.
https://pds-geosciences.wustl.edu/missions/mgs/mola.html
https://astrogeology.usgs.gov/search/details/Mars/GlobalSurveyor/MOLA/Mars_MGS_MOLA_DEM_mosaic_global_463m/cub

[2] Malin, M.C., Bell III, J.F., Cantor, B.A., Caplinger, M.A., Calvin, W.M., Clancy, R.T., Edgett, K.S., Edwards, L., Haberle, R.M., James, P.B., Lee, S.W., Ravine, M.A., Thomas, P.C., Wolff, M.J., 2007. Context camera investigation on board the Mars Reconnaissance Orbiter. J. Geophys. Res. 112, E05S04 https://doi.org/10.1029/ 2006JE002808.

[3] Dickson, J.L., Kerber, L.A., Fassett, C.I., Ehlmann, B.L., 2018. A global, blended CTX mosaic of Mars with vectorized seam mapping: a new mosaicking pipeline using principles of non-destructive image editing. In: 49th Lunar and Planetary Science Conference, 19–23 March, The Woodlands, TX, Abst. #2480.
http://murray-lab.caltech.edu/CTX/LPSC2018_CTX-Mosaic-Poster.pdf
http://murray-lab.caltech.edu/CTX/

[4] Fergason, R. L, Hare, T. M., & Laura, J. (2018). HRSC and MOLA Blended Digital Elevation Model at 200m v2. Astrogeology PDS Annex, U.S. Geological Survey. http://bit.ly/HRSC_MOLA_Blend_v0

[5] Jiang, C., Doute, S., Luo, B., & Zhang, L. Fusion of photogrammetric and photoclinometric information for high-resolution DEMs from Mars in-orbit imagery. ISPRS Journal of Photogrammetry and Remote Sensing. Volume 130, August 2017, Pages 418-430.
https://doi.org/10.1016/j.isprsjprs.2017.06.010

[6] Leclerc, Y.G., Bobick, A.F., 1991. The direct computation of height from shading, Computer Vision and Pattern Recognition, 1991. In: Proceedings CVPR'91., IEEE Computer Society Conference on. IEEE, pp. 552–558.

[7] Hess, M., Wohlfarth, K., Grumpe, A., Wöhler, C., Ruesch, O., and Wu, B.: ATMOSPHERICALLY COMPENSATED SHAPE FROM SHADING ON THE MARTIAN SURFACE: TOWARDS THE PERFECT DIGITAL TERRAIN MODEL OF MARS, Int. Arch. Photogramm. Remote Sens. Spatial Inf. Sci., XLII-2/W13, 1405–1411, https://doi.org/10.5194/isprs-archives-XLII-2-W13-1405-2019, 2019.

[8] Emmert-Streib F, Yang Z, Feng H, Tripathi S and Dehmer M (2020) An Introductory Review of Deep Learning for Prediction Models With Big Data. *Front. Artif. Intell.* 3:4. doi: 10.3389/frai.2020.00004

[9] Chan, T. H., Jia, K., Gao, S., Lu, J., Zeng, Z., & Ma, Y. (2015). PCANet: A simple deep learning baseline for image classification?. *IEEE transactions on image processing*, *24*(12), 5017-5032.

[10] Silver, D., Schrittwieser, J., Simonyan, K. *et al.* Mastering the game of Go without human knowledge. *Nature* **550,** 354–359 (2017). https://doi.org/10.1038/nature24270

[11] Qu, Y., Chen, Y., Huang, J., & Xie, Y. (2019). Enhanced pix2pix dehazing network. In *Proceedings of the IEEE Conference on Computer Vision and Pattern Recognition* (pp. 8160-8168).

[12] Jason Antic. 2019. Deoldify. (2019). https://github.com/jantic/DeOldify

[13] https://www.youtube.com/c/CtrlShiftFace/ https://www.digitaltrends.com/cool-tech/ctrl-shift-face-deepfake-changing-hollywood-history/

[14] https://unity.com/

[15] https://www.unrealengine.com/en-US/

[16] https://ode.rsl.wustl.edu/mars/productPageAtlas.aspx?product_idgeo=17984491